\documentclass{article}
\usepackage[utf8]{inputenc}
\usepackage{graphicx}
\usepackage{color}
\usepackage{float}
\usepackage{todonotes}
\usepackage{amsfonts}
\usepackage{amsthm}
\usepackage{amsmath}
\usepackage{caption}
\usepackage{amssymb}
\usepackage{stmaryrd}
\usepackage{mathtools}
\usepackage[top=2.5cm, bottom=2.5cm, left=3cm, right=3cm]{geometry}
\usepackage[english]{babel}
\usepackage[T1]{fontenc}
\usepackage{hyperref}
\usepackage[ruled,vlined]{algorithm2e}

\usepackage{algorithmicx}
\usepackage{lineno}
\usepackage{float}
\usepackage{csquotes}
\usepackage{colortbl}
\usepackage{comment}

\newcommand{\R}{\mathbb{R}}

\newcommand{\N}{\mathbb{N}}
\newcommand{\E}{\mathbb{E}}

\def\Pc{{\cal P}}
\def\Lc{{\cal L}}

\def\Fc{{\cal F}}
\def\Hc{{\cal H}}

\def\Uc{{\cal U}}

\def\Lc{{\cal L}}

\def\Nc{{\cal N}}

\def\Jc{{\cal J}}
\def \trans{^{\scriptscriptstyle{\intercal}}}
\DeclareMathOperator*{\argmin}{arg\,min}

\def \trans{^{\scriptscriptstyle{\intercal}}}

\def \E{\mathbb{E}}
\def \F{\mathbb{F}}
\def \I{\mathbb{I}}

\def \P{\mathbb{P}}

\def\bomu{{\boldsymbol \mu}}

\def\argmax{\mathop{\rm argmax}}
\def\argmin{\mathop{\rm argmin}}
\def\argmin_#1{\underset{#1}{\mathrm{argmin\, }}}

\def \d{\mathrm{d}}

\def \b1{\bf{1}}

\def \mfa{\mathfrak{a}} 
\def \mra{\mathrm{a}} 
\def \mrJ{\mathrm{J}} 
\def \mrG{\mathrm{G}} 
 
\def \d{\mathrm{d}}

\def\beqs{\begin{eqnarray*}}
\def\enqs{\end{eqnarray*}}
\def\beq{\begin{eqnarray}}
\def\enq{\end{eqnarray}}

\newtheorem{Theorem}{Theorem}[section]

\newtheorem{Remark}[Theorem]{Remark}

\def\argmin_#1{\underset{#1}{\mathrm{argmin\, }}}
\def\argmax_#1{\underset{#1}{\mathrm{argmax\, }}}

\numberwithin{equation}{section} 

\mathtoolsset{showonlyrefs}

\usepackage[backend=bibtex, maxbibnames=5, style=numeric]{biblatex} 
\begin{document}

\title{Actor critic learning algorithms for mean-field control \\ with moment neural networks
\thanks{This work is supported by  FiME, Laboratoire de Finance des March\'es de l'Energie, and the ``Finance and Sustainable Development'' EDF - CACIB Chair. H. Pham is also supported by the BNP-Paribas Chair ``Futures of Quantitative Finance".}
}

\author{Huy\^en \sc{Pham}\footnote{LPSM,  Universit\'e Paris Cité,  \& FiME \sf \href{mailto:pham at lpsm.paris}{pham at lpsm.paris}} 
\and
Xavier \sc{Warin}
\footnote{EDF R\&D \& FiME \sf \href{mailto:xavier.warin at edf.fr}{xavier.warin at edf.fr}}}

\maketitle

\begin{abstract}
We develop a new policy gradient and actor-critic algorithm for solving mean-field control problems within a continuous time  reinforcement learning setting. 
Our approach leverages a  gradient-based representation of the value function, employing parametrized randomized policies.  
The learning for both the actor (policy) and critic (value function) is facilitated  by a  class of moment neural network functions on the Wasserstein space of probability measures, and the key feature is to sample directly trajectories of distributions.   
A central challenge addressed in this study pertains to the computational treatment of an operator specific to the mean-field framework. 
To illustrate the effectiveness of our methods, we provide a comprehensive set of numerical results. These encompass diverse examples, including multi-dimensional settings and nonlinear quadratic mean-field control problems 
with  controlled volatility. 
\end{abstract}


\vspace{5mm}


\section{Introduction}

This paper is concerned with  the numerical resolution of mean-field (a.k.a. McKean Vlasov) control in continuous time in a partially model-free reinforcement learning setting.  The dynamics of the controlled mean field stochastic differential equation  
on $\R^d$ is in the form 
\begin{align} \label{SDEX} 
\d X_t &= \; b(t,X_t,\P_{X_t},\alpha_t) \d t + \sigma(t,X_t,\P_{X_t},\alpha_t) \d W_t, 
\quad 0 \leq t \leq T, \; X_0 \sim \mu_0, 
\end{align}
where $W$ is a standard $d$-dimensional Brownian motion on a filtered probability space $(\Omega,\Fc,\F=(\Fc_t)_t,\P)$,  $\mu_0$  $\in$ $\Pc_2(\R^d)$, the Wasserstein space of square integrable probability measures, 
$\P_{X_t}$ denotes the marginal distribution of $X_t$ at time $t$,  and  the control process $\alpha$ is valued in $A$  $\subset$ $\R^p$.   The coefficients $b,\sigma$ are in the separable  form: 
 \begin{align} \label{structural} 
{\bf (SC)} \hspace{1.5cm}  b(t,x,\mu,a)  \; = \; \beta(t,x,\mu) +  C(t,a), \quad \quad  \sigma\sigma\trans(t,x,\mu,a) \; = \;  \Sigma(t,x,\mu) + \vartheta(t,a),
 \end{align} 
 for $(t,x,\mu,a)$ $\in$ $[0,T]\times\R^d\times\Pc_2(\R^d)\times\R^p$, where  $\beta$, $\Sigma$ depending on the state variable and its probability distribution  are unknown functions, while  the coefficients $C$, $\vartheta$ on the control are   known functions on $[0,T]\times\R^p$.

The expected total cost associated to a control $\alpha$ is given by 
\begin{align}
\label{eq:toMin}
\E \Big[ \int_0^T f(X_t,\P_{X_t},\alpha_t) \d t + g(X_T,\P_{X_T}) \Big], 
\end{align}
and we denote by $(t,x,\mu)$ $\mapsto$ $V(t,x,\mu)$ the associated value function defined on $[0,T]\times\R^d\times\Pc_2(\R^d)$. The analytical forms of $f$, $g$ are unknown, but it is  assumed that 
given an input $(x,\mu,a)$ $\in$ $\R^d\times\Pc_2(\R^d)\times A$, we obtain  (by observation or from a blackbox)  the realized output  costs $f(x,\mu,a)$ and $g(x,\mu)$.
Such setting  is  partially model-free, as the model coefficients $\beta$, $\Sigma$, $f$, and $g$ are unknown, and we only assume that the action functions $C$, $\vartheta$  on the control are known.

The theory and applications of mean-field control (MFC) problems  have  generated a vast literature in the last decade, and we refer to the  monographs \cite{cardel18}, \cite{cardel18b} for a comprehensive treatment of this topic. 
From a numerical aspect,  the main challenging issue is the infinite dimensional feature of MFC coming from the distribution law state variable. In a model-based setting, i.e., when all the coefficients $b$, $\sigma$, $f$ and $g$ are known, several recent works 
have proposed  deep learning schemes for MFC, based on neural network approximations of the feedback control and/or the value function solution to the Hamilton-Jacobi-Bellman equation from the dynamic programming or backward stochastic differential equations (BSDEs) from the maximum principle, see \cite{carlau22}, \cite{germain2022deepsets}, \cite{germikwar22}, \cite{hanhulong22}, \cite{Ruthotto9183}, \cite{reistozha21}, \cite{phawar22}.

The approximation of solutions to MFC in a (partially) model-free setting is the purpose of reinforcement learning (RL) where one learns in an unknown environment  
the optimal control (and the value function) by repeatedly trying  policies, observing state, receiving and 
evaluating rewards, and improving policies.  RL is a very active branch of machine learning, see the seminal reference monograph \cite{sutbar18}, and has recently attracted attention in the context of mean-field control in discrete-time, and mostly by $Q$-learning methods, see \cite{carlautan22}, \cite{angfoulau21}, \cite{guetal21}.

In this paper, we consider a partially model-free continuous time setting as described above, and adopt a policy gradient approach as in \cite{frietal23}. This relies on a gradient representation of the cost functional associated to randomized policies, which makes appear an additional operator term  $\Hc$ compared to the classical diffusion setting of \cite{jiazho21}, and specific to the  mean-field  setting.  The computational treatment of this operator $\Hc$ on functions defined on the Wasserstein space is the crucial issue, and has been  handled in \cite{frietal23} only for one-dimensional linear quadratic (LQ)  models, hence with a very particular  dependence of the value function and optimal control on the law of the state.  
Here, we address the general dependence of the coefficients on the distribution state, and deal with the operator $\Hc$ by means of the class of moment neural networks. This class of neural networks consists of  functions that depend on the measure via its first $L$ moments, and  satisfies  some  universal approximation theorem for functions defined on the Wasserstein space, see \cite{pham2022mean}, \cite{warin2023quantile}.  
We then design an actor-critic algorithm for learning alternately the optimal policy (actor) and value function (critic) with moment neural networks, 
which provides an effective resolution  of MFC control problems in a (partially) model-free setting beyond the LQ setting for multivariate dynamics with control on the drift and the volatility.  Our actor-critic algorithm has the structure of general actor-critic algorithms but during gradient iterations, instead of following a single state 
trajectory by sampling, we follow the evolution of  an entire distribution that is initially randomly chosen and
 described by its empirical measure obtained with a large fixed number of particles. Then, the batch
version of the algorithm consists in sampling and following $m$ distributions together to estimate the gradients. 

The outline of the paper is organized as follows.  We recall in Section \ref{sec:preli} the gradient representation of the functional cost with randomized policies and formulate notably the expression of the operator $\Hc$. 
In Section \ref{sec:param},  we consider the class of moment neural networks, and show how it acts on the operator $\Hc$.  We present in Section \ref{sec:algo} the actor-critic algorithm, and  
Section \ref{sec:num} is devoted to numerical results for illustrating the accuracy and efficiency of our algorithm. We present various examples with control on the drift and on the volatility, and non LQ examples in a multi-dimensional setting.


\vspace{1mm}

\noindent {\bf Notations.} The scalar product between two vectors $x$ and $y$ is denoted by $ x\cdot y$, and $|\cdot|$ is the Euclidian norm.  Let $Q$ $=$ $(Q_{i_1\ldots i_q})$  $\in$ $\R^{d_1\times \ldots\times d_q}$ be a tensor of order $q$, and 
$P$ $=$ $(P_{i_1\ldots i_p})$  $\in$ $\R^{d_1\times \ldots\times d_p}$ be a tensor of order $p$ $\leq$ $q$.   We denote by $Q \circ P$ the circ product defined as the tensor in $\R^{d_{p+1}\times \ldots \times d_q}$ with components: 
\begin{align}
[Q \circ P]_{i_{p+1} \ldots i_q} &= \; \sum_{i_1,\ldots,i_p} Q_{i_1\ldots i_p i_{p+1}\ldots i_q} P_{i_1\ldots i_p}. 
\end{align}
When $q$ $=$ $p$ $=$ $1$,  $\circ$ is the scalar product in  $\R^{d_1}$.  When $q$ $=$ $2$, $p$ $=$ $1$,  $Q\circ P$ $=$ $Q\trans P$ $\in$ $\R^{d_2}$ where $\trans$ is the transpose matrice operator. When $q$ $=$ $p$ $=$ $2$, 
$\circ$ is the inner product $Q\circ P$ $=$ ${\rm tr}(Q\trans P)$ where ${\rm tr}$ is the trace operator.  When $q$ $=$ $3$, $Q\circ P$ is a vector in $\R^{d_3}$ for $p$ $=$ $2$, and a matrix in $\R^{d_2\times d_3}$ for $p$ $=$ $1$.

\section{
Preliminaries
} \label{sec:preli}

 We  adopt a policy gradient  approach by searching optimal control among  parametrized randomized policies, i.e., family of probability transition kernels $\pi_\theta$ from $[0,T]\times\R^d\times\Pc_2(\R^d)$ into $\R^p$, with densities $p_\theta(t,x,\mu,.)$ w.r.t. some measure on $\R^p$, 
 and thus  by minimizing over the parameters $\theta$ $\in$ $\R^D$  the functional 
 \begin{align} \label{defJ}
 \mrJ(\theta) & = \;  \E_{\alpha\sim\pi_\theta} \Big[ \int_0^T f(X_t,\P_{X_t},\alpha_t) 
 \d t + g(X_T,\P_{X_T}) \Big].
 \end{align} 
 Here $\alpha$ $\sim$ $\pi_\theta$ means that at each time $t$, the action $\alpha_t$ is sampled (independently from $W$) from the probability distribution $\pi_\theta(.|t,X_t,\P_{X_t})$. 

  \begin{Remark}
 We may include  a entropy (e.g. Shannon) regularizer term in the functional cost \eqref{defJ} as proposed in \cite{wang2020reinforcement} for encouraging exploration of randomized policies.  
 This can slightly help the convergence of the policy gradient algorithms by permitting the use of higher learning rates, but it turns out that it does not really improve the accuracy of the results. Here, we only consider exploration through the randomization of policies. 
 \end{Remark}

 
 \vspace{1mm}

We have the gradient representation of $\mrJ$ as derived  in \cite{frietal23}:  
 \begin{align} 
\mrG(\theta) &:= \; \nabla_\theta \mrJ(\theta)  \\
 & = \;  \E_{\alpha\sim\pi_\theta} \Big[ \int_0^T \nabla_\theta \log  p_\theta(t,X_t,\P_{X_t},\alpha_t) \big[ \d J_\theta(t,X_t,\P_{X_t}) + 
 f(X_t,\P_{X_t},\alpha_t) 
 \d t  \big] \\
 & \hspace{3cm}  + \; \int_0^T \Hc_\theta[J_\theta](t,X_t,\P_{X_t}) \d t \Big],  \label{grad} 
 \end{align} 
where $J_\theta$ $:$ $[0,T]\times\R^d\times\Pc_2(\R^d)$ $\rightarrow$ $\R$  is the dynamic value function  associated to \eqref{defJ}, hence satisfying the property that 
\begin{align} \label{marJ} 
{\bf (MJ)} \quad \{J_\theta(t,X_t,\P_{X_t})  + \int_0^t 
f(X_s,\P_{X_s},\alpha_s) 
\d s, \; 0 \leq t\leq T\}  \;    \mbox{ is a martingale}, 
\end{align} 
and $\Hc_\theta$ is the operator specific to the mean-field framework, defined by 
\begin{align}
\Hc_\theta[\varphi](t,x,\mu) &= \;  
 \nabla_\theta \E_{\xi\sim\mu} \Big[   b_\theta (t,\xi,\mu) \cdot \partial_\mu \varphi(t,x,\mu)(\xi) + \frac{1}{2}      \Sigma_\theta(t,\xi,\mu) \circ   \partial_\xi \partial_\mu \varphi(t,x,\mu)(\xi) \Big] \; \in \R^D
\end{align}
with $b_\theta(t,x,\mu)$ $=$ $\int_A b(t,x,\mu,a) \pi_\theta(\d a |t,x,\mu)$, $\Sigma_\theta(t,x,\mu)$ $=$ $\int_A \sigma\sigma\trans(t,x,\mu,a)  \pi_\theta(\d a |t,x,\mu)$. 
Here  $\partial_\mu \varphi(t,x,\mu)(.)$ is the Lions-derivative with respect to $\mu$ $\in$ $\Pc_2(\R^d)$, and it is a function from  $\R^d$ into $\R^d$, and $\E_{\xi\sim\mu}[.]$ means that the expectation is taken with respect to the random variable $\xi$ distributed according to $\mu$.


Notice that under the structure condition {\bf (SC)}, we have 
\beqs
b_\theta(t,x,\mu) \; = \; \beta(t,x,\mu) + C_\theta(t,x,\mu), 
\quad \Sigma_\theta(t,x,\mu) \;  = \;  \Sigma(t,x,\mu) + \vartheta_\theta(t,x,\mu)
\enqs
where $C_\theta(t,x,\mu)$ $:=$ $\int_A C(t,a) \pi_\theta(\d a |t,x,\mu)$, $\vartheta_\theta(t,x,\mu)$ $:=$ $\int_A \vartheta(t,a)  \pi_\theta(\d a |t,x,\mu)$ are known functions, and thus 
\begin{align}
\Hc_\theta[\varphi](t,x,\mu) 
& = \;  \nabla_\theta\E_{\xi\sim\mu} \Big[  C_\theta (t,\xi,\mu) \cdot  \partial_\mu \varphi(t,x,\mu)(\xi) + \frac{1}{2}  \vartheta_\theta(t,\xi,\mu) \circ   \partial_\xi \partial_\mu \varphi(t,x,\mu)(\xi) \Big].  
\end{align}

\section{Parametrization of actor/critic functions with moment neural networks} \label{sec:param}

A moment neural network function on $[0,T]\times\R^d\times\Pc_2(\R^d)$ of order $L$ $\in$ $\N^*$  is a parametric function in the form 
\begin{align}
\phi_\eta(t,x,\mu) &= \;    \Psi_\eta(t,x,\bar\bomu_L),
\end{align}
where  $\bar\bomu_L$ $=$ $(\bar\mu^\ell)_{\ell \in \Lc}$, with  $\bar\mu^\ell$ $=$ $\E_{\xi\sim\mu}[\prod_{i=1}^d \xi_i^{\ell_i}]$ for $\ell$ $=$ $(\ell_i)_{i \in \llbracket 1,d\rrbracket}$ $\in$ $\Lc$ $=$ $\{ \ell = (\ell_1,\ldots,\ell_d) \in \N^d: \sum_{i=1}^d \ell_i \leq L\}$ of cardinality $L_d$, and  
$(t,x,y)$ $\in$ $[0,T]\times\R^d\times\R^{L_d}$ $\mapsto$ $\Psi_\eta(t,x,y)$  
is a classical finite-dimensional feedforward neural network with parameters $\eta$. Moment neural networks have been considered  in \cite{warin2023quantile} as a 
special case of cylindrical mean-field neural networks, and  satisfy a  universal approximation theorem for continuous functions  on $[0,T]\times\R^d\times\Pc_2(\R^d)$, see \cite{pham2022mean}. 
By abuse of notation and language, we identify $\phi_\eta$ and $\Psi_\eta$, and call them  indifferently moment neural networks.

We shall parametrize  the randomized policy (actor) by a Gaussian probability transition kernel in the form
\beqs
\pi_\theta(.|t,x,\mu) &=& \Nc( m_\theta(t,x,\bar\bomu_L), \lambda \I_p), 
\enqs
where $m_\theta$ is a moment neural network,  hence with log density:
\beqs
\log p_\theta(t,x,\mu,a) &=& - \frac{1}{2} \log (2\pi \lambda) - \frac{ |a-m_\theta(t,x,\bar\bomu_L)|^2}{2\lambda},
\enqs
and $\lambda$ $>$ $0$ is a parameter for exploration. 
Notice that in this case,  the known functions $C_\theta$, $\vartheta_\theta$ depend on $\mu$ only though its $L$ moments $\bar\bomu_L$, and by misuse of notation we also write: $C_\theta(t,x,\bar\bomu_L)$, $\vartheta_\theta(t,x,\bar\bomu_L)$.

The value function (critic) is parametrized by a moment neural network  $\Jc_\eta(t,x,\mu)$ $=$ $\Jc_\eta(t,x,\bar\bomu_L)$, and we notice that
\begin{align}
\partial_\mu \Jc_\eta(t,x,\mu)(\xi) &= \;  D_1(\xi) \circ  \nabla_y \Jc_\eta(t,x,\bar\bomu_L) \\
\partial_\xi \partial_\mu \Jc_\eta(t,x,\mu)(\xi      
) &= \;  D_2(\xi) \circ  \nabla_y \Jc_\eta(t,x,\bar\bomu_L),
\end{align}
where $D_1(\xi)$ is  the matrix in $\R^{L_d\times d}$, and  $D_2(\xi)$ is the tensor in $\R^{L_d\times d\times d}$ with components 
\begin{align}
[D_1(\xi)]_{\ell i} &= \;  \ell_i\xi_i^{\ell_i-1}\prod_{k\neq i} \xi_k^{\ell_k}, \quad \mbox{ for } \xi = (\xi_i)_{i\in\llbracket 1,d\rrbracket}, \; \ell = (\ell_i)_{i\in\llbracket 1,d\rrbracket},  \\
[D_2(\xi)]_{\ell ij} &= \; \begin{cases}
				\ell_i(\ell_i-1)\xi_i^{\ell_i-2}\prod_{k\neq i} \xi_k^{\ell_k}, \quad i =j \\
				\ell_i \ell_j \xi_i^{\ell_i-1}\xi_j^{\ell_j-1} \prod_{k\neq i,j} \xi_k^{\ell_k}, \quad i \neq j.
				\end{cases}	
\end{align}

The expression of the operator $\Hc_\theta$ applied to the moment neural network critic function is then given by 
\begin{align}  
\Hc_\theta[\Jc_\eta](t,x,\mu) &= \;   \nabla_\theta \Big[ \E_{\xi\sim\mu} \big[  D_1(\xi) C_\theta (t,\xi,\bar\bomu_L)  + \frac{1}{2} D_2\trans(\xi) \circ  \vartheta_\theta(t,\xi,\bar\bomu_L) \big] \cdot     \nabla_y \Jc_\eta(t,x,\bar\bomu_L) \Big].  
\label{expressH} 
\end{align}
Here $D_2\trans(\xi)$ is the tensor in $\R^{d\times d\times L_d}$ with components $[D_2\trans(\xi)]_{ij\ell}$ $=$ $[D_2(\xi)]_{\ell ij}$. 

In the algorithm, we shall use the expectation of $\Hc_\theta$, which is given from \eqref{expressH}  by 
\begin{align} \label{espH} 
\overline\Hc_\theta[\Jc_\eta](t,\mu) & := \; \E_{\xi \sim \mu} \big[ \Hc_\theta[\Jc_\eta](t,\xi,\mu) \big] \\
&= \;  
\nabla_\theta \Big[ \E_{\xi\sim\mu} \big[  D_1(\xi) C_\theta (t,\xi,\bar\bomu_L)  + \frac{1}{2} D_2\trans(\xi) \circ  \vartheta_\theta(t,\xi,\bar\bomu_L) \big] \cdot     \nabla_y  \E_{\xi\sim\mu} \big[ \Jc_\eta(t,\xi,\bar\bomu_L) \big]  \Big].
\end{align}

\begin{Remark} \label{remH} 
{\bf 1.} For complexity argument, it is crucial to rely on the above expression of the operator $\Hc_\theta$ where the  differentiation is taken on the expectation $\E_{\xi\sim\mu}[.]$, and not the reversal:  expectation of the differentiation. 
Indeed, in the latter case, after empirical approximation of the expectation with $M$ samples $\xi^j$ $\sim$ $\mu$, $j$ $=$ $1,\ldots,M$,  one should compute by automatic differentiation 
\beqs
\nabla_\theta \big[ D_1(\xi^j) C_\theta (t,\xi^j,\bar\bomu_L)  + \frac{1}{2} D_2\trans(\xi^j) \circ  \vartheta_\theta(t,\xi^j,\bar\bomu_L)  \big] , \quad \nabla_y  \Jc_\eta(t,\xi^j,\bar\bomu_L), \quad j=1,\ldots,M,  
\enqs
which is very costly as $M$ is of order $10^4$. In the former case, $\overline\Hc_\theta[\Jc_\eta](t,\mu)$ is approximated  by automatic differentiation via
\beqs
\widehat \Hc_\theta^M[\Jc_\eta](t,\mu) & := \;  \nabla_\theta  \Big[ \frac{1}{M} \sum_{j=1}^M  D_1(\xi^j) C_\theta (t,\xi^j,\bar\bomu_L)  + \frac{1}{2} D_2\trans(\xi^j) \circ  \vartheta_\theta(t,\xi^j,\bar\bomu_L)  \cdot     \nabla_y \frac{1}{M} \sum_{j=1}^M  \Jc_\eta(t,\xi^j,\bar\bomu_L) \Big] .
\enqs
hence saving an order $M$ for the complexity cost.  
In theory, it is also possible to choose other networks for taking into account the dependency on the distribution $\mu$: the cylindrical network proposed in \cite{pham2022mean} could be used but some automatic differentiation are then requested to calculate $\partial_\mu \Jc_\eta(t,x,\mu)(\xi)$ and  $\partial_\xi \partial_\mu \Jc_\eta(t,x,\mu)(\xi)$ for each sample of $\xi$ leading to an explosion in the computation time. 
\vspace{1mm}

\noindent {\bf 2.}  In order to calculate the term $\nabla_y  \Jc_\eta$, it is necessary to explore different initial distributions, otherwise $\Jc_\eta$ only depends on $t$ and $x$ at convergence and the gradient is impossible to estimate.
\end{Remark}


\section{Algorithm} \label{sec:algo} 

The actor-critic  method consists in two optimization stages that are  performed alternately: 
\begin{itemize}
\item[(1)] 
{\it Policy evaluation}:  given an actor  policy $\pi_\theta$, evaluate  its cost functional with the critic function $\Jc_\eta$ that minimizes the loss function arising from the martingale property {\bf (MJ)} after time discretization of the interval $[0,T]$ with the time grid $\{t_k= k \Delta t, k=0,\ldots,n\}$:
\begin{align}
L^{PE}(\eta) &= \;  \E_{\alpha\sim\pi_\theta} \Big[ \sum_{k=0}^{n-1} \big| g_{t_n}  + \sum_{l=k}^{n-1} 
f_{t_l} 
\Delta t - \Jc_\eta(t_k,X_{t_k},\mu_{t_k}) \big|^2 \Delta t \Big], 
\end{align}
where we set $\mu_{t_l}$ $=$ $\P_{X_{t_l}}$ for the law of $X_{t_l}$,  and $f_{t_l}$ $=$ $f(X_{t_l},\mu_{t_l},\alpha_{t_l})$ as the output cost at time $t_l$ for input state $X_{t_l}$, law $\mu_{t_l}$, action $\alpha_{t_l}$ $\sim$ $\pi_\theta(.|t_l,X_{t_l},\mu_{t_l})$, and $g_{t_n}$ $=$ 
$g(X_{t_n},\mu_{t_n})$ the terminal output cost for input $X_{t_n}$, $\mu_{t_n}$. 
\item[(2)] {\it Policy gradient}: given a critic cost function $\Jc_\eta$, update the parameter $\theta$ of the actor by stochastic gradient descent by using the gradient, which is given from \eqref{grad} and after time discretization by 
\begin{align}
G(\theta) &= \; \E_{\alpha\sim\pi_{\theta}} \Big[  \sum_{k=0}^{n-1} \nabla_\theta \log p_\theta(t_k,X_{t_k},\mu_{t_k},\alpha_{t_k}) \big[ \Jc_\eta(t_{k+1},X_{t_{k+1}},\mu_{t_{k+1}}) - \Jc_\eta(t_k,X_{t_k},\mu_{t_k})  \\
& \hspace{3cm}  + \; 
f_{t_k} 
\Delta t \big]  + \Hc_\theta[\Jc_\eta](t_k,X_{t_k},\mu_{t_k}) \Big]. 
\end{align}
\end{itemize}

In the practical implementation, we proceed as follows  for each epoch  e (gradient iteration descent) with  a given exploration parameter  $\lambda(e)$ decreasing to 0: 
\begin{itemize}
\item We start with a batch $N$ (of order $10$) of initial distributions $\mu_0^i$, $i$ $=$ $1,\ldots,N$, e.g. Gaussian distributions by varying the mean and std deviations parameters, and sample $X_{0}^{i,j}$ $\sim$ $\mu_0^i$, $j$ $=$ $1,\ldots,M$ with $M$ of order $10^4$. If our ultimate goal is to learn the optimal control and function value for other families of initial distributions, the initial distributions should be sampled accordingly.
\item We then run by forward induction in time: for $k$ $=$ $0,\ldots,n-1$:
\begin{itemize}
\item[-]  Empirical estimate of  $\mu_{t_k}^i$ from $(X^{i,j}_{t_k})_{j\in \llbracket 1, M\rrbracket}$, for $i$ $=$ $1,\ldots,N$.  
\item[-] Sample $\alpha_{t_k}^{i,j}$ $\sim$ $\pi_\theta(.|t_k,X_{t_k}^{i,j},\mu_{t_k}^i)$, $i$ $\in$ $\llbracket 1,N\rrbracket$, $j$ $\in$ $\llbracket 1,M\rrbracket$
using the exploration parameter $\lambda(e)$
\item[-] Observe running cost $f_{t_k}^{i,j}$ $=$ $f(X_{t_k}^{i,j},\mu_{t_k}^i,\alpha_{t_k}^{i,j})$, and next state $X_{t_{k+1}}^{i,j}$,  $i$ $\in$ $\llbracket 1,N\rrbracket$, $j$ $\in$ $\llbracket 1,M\rrbracket$
\end{itemize}
\item Observe final cost $g_{t_n}^{i,j}$ $=$ $g(X_{t_n}^{i,j},\mu^i_{t_n})$, $i$ $\in$ $\llbracket 1,N\rrbracket$, $j$ $\in$ $\llbracket 1,M\rrbracket$
\item  
Compute the empirical mean approximation of $L^{PE}(\eta)$ on all  initial distributions $\mu_0^i$, $i$ $\in$ $\llbracket 1,N\rrbracket$:  
\begin{align}
\widetilde L_M^{PE}(\eta) &= \;  \frac{1}{M N}  \sum_{i=1}^N \sum_{j=1}^M \sum_{k=0}^{n-1} \big| g_{t_n}^{i,j}  + \sum_{l=k}^{n-1} 
f_{t_l}^{i,j} 
\Delta t - \Jc_\eta(t_k,X^{i,j}_{t_k},\mu^i_{t_k}) \big|^2 \Delta t, 
\end{align}
and update the critic parameter by 
\begin{align}
\eta  &\longleftarrow \; \eta  -  \rho^C \nabla_\eta \tilde L_M^{PE}(\eta),
\end{align}
where $\rho^C$ is a learning rate. Notice that the gradient is calculated by automatic differentiation. 
\item 
Compute the empirical mean approximation  of   $G(\theta)$ on all  initial distributions $\mu_0^i$, $i$ $\in$ $\llbracket 1,N\rrbracket$:
\begin{align}
 \widetilde G_M(\theta) &= \;  \nabla_\theta  \frac{1}{M N} \sum_{i=1}^N   \sum_{j=1}^M  \Big\{  \sum_{k=0}^{n-1}  \log p_\theta(t_k,X^{i,j}_{t_k},\mu^i_{t_k},\alpha^{i,j}_{t_k}) \big[ \Jc_\eta(t_{k+1},X^{i,j}_{t_{k+1}},\mu^i_{t_{k+1}}) - \Jc_\eta(t_k,X^{i,j}_{t_k},\mu^i_{t_k})  \\
& \hspace{3cm}  + \; 
f^{i,j}_{t_k} 
\Delta t \big]  \Big\}
+    \widetilde \Hc^M_\theta[\Jc_\eta]
\end{align}
where 
\begin{align*}
  \widetilde \Hc^M_\theta[\Jc_\eta]  & = &  \nabla_\theta  \Big (  \frac{1}{N} \sum_{i=1}^N   \Big[  \sum_{k=0}^{n-1}  
\Big(  \frac{1}{M} \sum_{j=1}^M   D_1(X_{t_k}^{i,j})  C_\theta (t_k,X_{t_k}^{i,j},\mu_{t_k}^i) 
+ \frac{1}{2} D_2\trans (X_{t_k}^{i,j}) \circ  \vartheta_\theta(t_k,X_{t_k}^{i,j},\mu_{t_k}^i) \Big)  \cdot \\
& & \hspace{2cm} \nabla_y \Big(  \frac{1}{M} \sum_{j=1}^M   \Jc_\eta(t_k,X_{t_k}^{i,j},\mu_{t_k}^i) \Big)  \Big] \Big)
\end{align*}
and update the actor parameter by
\begin{align}
\theta  &\longleftarrow \; \theta  -  \rho^A   \widetilde G_M(\theta)  ,
\end{align}
where $\rho^A$ is a learning rate.  Again for efficiency, it is crucial to compute by automatic differentiation the gradient after computing all the different  expectations as in  Remark \ref{remH}. 
\end{itemize}
The output  $(\theta^*, \eta^*)$ are  the optimal  parameters obtained at convergence of the algorithm. 

\begin{Remark}
Compared to classical actor critic algorithm where one samples a trajectory for a given distribution, here the batch version of the algorithm consists in sampling and following $N$ distributions together to estimate the gradients.
\end{Remark}

\begin{Remark}
In order to check that the algorithm has effectively converged to the solution, we can use the calculated control $m_{\theta^*}(t,x,\mu)$ and apply it from  different initial distributions  $\mu_0$ sampled as $(X_0^j)_{j \in\llbracket 1,M\rrbracket}$ in a time discretized version of \eqref{eq:toMin}. Taking discrete expectation, we can compare the result obtained to $\frac{1}{M} \sum_{j=1}^M \Jc_{\eta^*}(0, X_0^j, \mu_0)$. When results are very close, we can suppose that the algorithm has effectively converged to the right solution.
\end{Remark}

\begin{Remark}
In the case where we know a priori that the running cost and terminal cost functions depend on the probability distribution $\mu$ only via its moments $\bar\bomu_L$, then we only need to estimate the moments of  $\mu_{t_k}^i$ from $(X^{i,j}_{t_k})_{j\in \llbracket 1, M\rrbracket}$, since all the other coefficients in the algorithm depend upon  the measure via its moments. 
\end{Remark}

 \begin{Remark}
     When $C_\theta(t,x,\mu)$ $:=$ $\int_A C(t,a) \pi_\theta(\d a |t,x,\mu)$, $\vartheta_\theta(t,x,\mu)$ $:=$ $\int_A \vartheta(t,a)  \pi_\theta(\d a |t,x,\mu)$ are not analytically explicit, it is always possible to estimate them numerically for example using a  numerical quadrature or a quasi Monte carlo/Monte-Carlo method but with some non negligible extra costs.
 \end{Remark}

\section{Numerical results} \label{sec:num} 

\begin{figure}[H]
\centering
\includegraphics[width=0.3\textwidth]{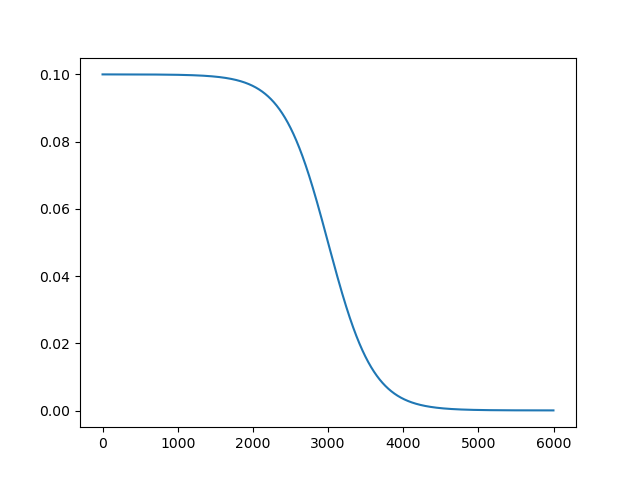}
\caption{Exploration function $\lambda$ for a number of epochs $\hat N= 6000$. \label{fig:lamFig}}
\end{figure}

Throughout this section, we use  moment neural networks  with 3 hidden layers and 20 neurons on each layer, and choose the activation function $\tanh$. 
The exploration parameter  $\lambda$  is a function of the number of gradient descent iterations  (epoch number $e$ $\leq$  $\hat N$): 
\begin{align}
\label{eq:lam}
    \lambda(e)= (\bar \lambda- \underline \lambda)\Big(1- S\big(\frac{20 e -10 \hat N }{ \hat N}\big)  \Big) + \underline \lambda, 
\end{align}
where $\underline \lambda= 0.0001$ and $\bar \lambda =0.1$ and $S$ is the sigmoid function: $S(x)= \frac{1}{1+ \exp(-x)}$. In other words, it  is chosen so that the exploration period with $\lambda$ close to $0.1$ is long enough, then $\lambda$ slowly decreases to $0.0001$ and stays close to that value long enough. This fonction is plotted on Figure \ref{fig:lamFig}.

At a gradient descent iteration $e$, we sample the control as:
\begin{align*}
    \pi_\theta(.|t, x, \mu) \sim {\mathcal N}( m_\theta(t,x, \mu), \lambda(e) \I_p).
\end{align*}
During the gradient descent algorithm,  we use the ADAM optimizer \cite{kingma2014adam}. We point out that it is crucial to use  two timescales approach (see \cite{angfoulau21}), for the  learning rates $\rho^C$, $\rho^A$  of  the critic and actor updates:  
$\rho^C$ should be at least one order of magnitude higher than $ \rho^A$ to get good convergence, hence  the approximate critic function  should evolve faster. 
We take a batch size $N=10$ while the number of samples to estimate distributions is taken equal to $M =10000$ or $M=20000$ depending on the examples.

In the tables and figures below,  we give the average analytic solution "Anal" at $t=0$, i.e. $\E_{X_0 \sim \mu_0}[V(0,X_0,\mu_0)]$, and  the average calculated value function "Calc":  $\E_{X_0 \sim \mu_0}[\Jc_{\eta^*}(0,X_0,\mu_0)]$  obtained by the algorithm at $t=0$, by varying the  initial distributions $\mu_0$.  
The MSE is the mean square error between the analytic and the critic value  computed at $t=0$, i.e. $ \E_{X_0 \sim \mu_0}\big|\Jc_{\eta^*}(0,X_0,\mu_0) - V(0,X_0,\mu_0)\big|^2$, and the relative error is 
\begin{align}
{\rm RelError} &= \; \frac{ \E_{X_0\sim\mu_0}\big[ \Jc_{\eta^*}(0,X_0,\mu_0) - V(0,X_0,\mu_0)\big]}{ \E_{X_0\sim\mu_0}[V(0,X_0,\mu_0)]} 
\end{align}
We shall also plot in the one-dimensional case $d$ $=$ $1$,  and $A$ $=$ $\R$, 
the trajectories of the optimal control $t$ $\mapsto$ $\alpha_t^*$ vs the ones obtained from moment neural networks, i.e., 
$t$ $\mapsto$ $m_{\theta^*}(t,X_t,(\E[X_t^\ell])_{\ell \in \llbracket 1,L\rrbracket})$.

All training times are calculated on a on GPU NVidia V100 32Go  graphic card.

We consider four examples with control on the drift,  including multidimensional setting and nonlinear quadratic mean-field control, and one example with controlled volatility, for which we have analytic solutions to be compared with the approximations calculated from our actor critic algorithm.

\subsection{Examples with controlled drift}

In the four examples of  this paragraph, we take $\vartheta(t,a)$ $\equiv$ $0$, $C(t,a)$ $=$ $a$ and so  $\vartheta_\theta$ $\equiv$ $0$, $C_\theta$ $=$ $m_\theta$.  

\subsubsection{Systemic risk model in one dimension} \label{sec:sys1D}

We consider the model in \cite{carfousun15}:

\begin{equation}  \label{modelsys1}
\begin{cases}
b(x,\mu,a) &= \; \kappa(\bar\mu - x) + a, \quad  \sigma \mbox{ positive constant }  \\
f(x,\mu,a)  &  = \; \frac{1}{2}a^2 - qa(\bar \mu-  x) + \frac{p}{2} (\bar \mu-x)^2, \quad  \quad  g(x,\mu) \; = \; \frac{c}{2}(x-\bar \mu)^2,
\end{cases}
\end{equation}
for $(x,\mu,a)$ $\in$ $\R\times\Pc_2(\R)\times\R$, with some positive constants $\kappa$, $q$, $p$, $c$ $>$ $0$,  $q^2$ $\leq$ $p$. Here we denote by  $\bar \mu$  $:=$ $\E_{\xi \sim \mu}[\xi]$.

In this linear quadratic (LQ) model, the value function is explicitly  given by
\begin{align}
V(t,x, \mu) &= \; K(t) (x- \bar \mu)^2 + \sigma^2 R(t), \label{eq:sys1D}
\end{align}
where 
\begin{align}
K(t) & = 
- \frac{1}{2} \Big[ \kappa + q - \sqrt{\Delta} \frac{ \sqrt{\Delta} \sinh(\sqrt{\Delta}(T-t))  + (\kappa + q + c) \cosh (\sqrt{\Delta}(T-t))}{ \sqrt{\Delta} \cosh(\sqrt{\Delta}(T-t))  + 
(\kappa + q + c) \sinh (\sqrt{\Delta}(T-t))} \Big],   
\end{align}     
with $\sqrt{\Delta}$ $=$ $\sqrt{(\kappa+q)^2  + p -  q^2}$,  and 
\begin{align}
R(t) & = \;  \frac{\sigma^2}{2} \ln \Big[ \cosh (\sqrt{\Delta}(T-t)) + \frac{\kappa + q  + c}{\sqrt{\Delta}}  \sinh (\sqrt{\Delta}(T-t)) \Big] - \frac{\sigma^2}{2} (\kappa + q)(T-t),
\end{align} 
while the optimal control is given by 
\begin{align}
\alpha_t^* &= \;  (2 K(t) + q)(\E[X_t] - X_t), \quad 0 \leq t \leq T.   
\end{align}

The parameters of the model are fixed to the following values: $\kappa =0.6$ , $\sigma= 1$, $p= c=  2$, $q=0.8$, $T=1$. We take a number of time steps $n=100$,  $M= 10000$.
At each gradient iteration, the initial distribution is sampled with
\begin{align*}
X_0 & \sim \; \mu_0  \; = \;    \upsilon_0 \; {\mathcal N}(0,1), 
\end{align*}
where $\upsilon_0^2$ is sampled at each iteration according to the uniform distribution on $[0,1]$ for each element of the batch. 
In the table, we give the results obtained in simulation with $\upsilon_0^2 \in \{0, \frac{1}{10}, \ldots, 1\}$ after training with $L=2$ moments, using $\hat N = 6000$ gradient iterations. 

\begin{table}[H]
 \centering
    \begin{tabular}{|c|c|c|c|c|c|} \hline 
       $\upsilon_0^2$  &  0. & 0.1 & 0.2 & 0.3 & 0.4  \\ \hline 
       Anal & 0.3870 &0.4095 &0.4321 &0.4546 &0.4772  \\ \hline
Calc
& 0.3958 &0.4198 &0.4421 &0.4642 &0.4875 \\ \hline
MSE & 0.0001 &0.0001 &0.0002 &0.0002 &0.0004  \\ \hline

    \end{tabular} 
\begin{tabular}{|c|c|c|c|c|c|c|} \hline 
       $\upsilon_0^2$  &   0.5 & 0.6 & 0.7 &0.8 & 0.9 \\ \hline 
       Anal & 0.4997 &0.5223 &0.5448 &0.5674 &0.5900 \\ \hline
Calc
& 0.5112 &0.5341 &0.5593 &0.5858 &0.6082 \\ \hline
MSE & 0.0005 &0.0005 &0.0006 &0.0005 &0.0007 \\ \hline
    \end{tabular}
    \caption{ \footnotesize{Results for the systemic model using $L=2$, $\rho^A=0.0005$, $\rho^C=0.01$. Training time is 106863s}.
    \label{tab:sysMom2}}
\end{table}

\begin{figure}[H]
     \centering
 \begin{minipage}[t]{0.32\linewidth}
  \centering
 \includegraphics[width=\textwidth]{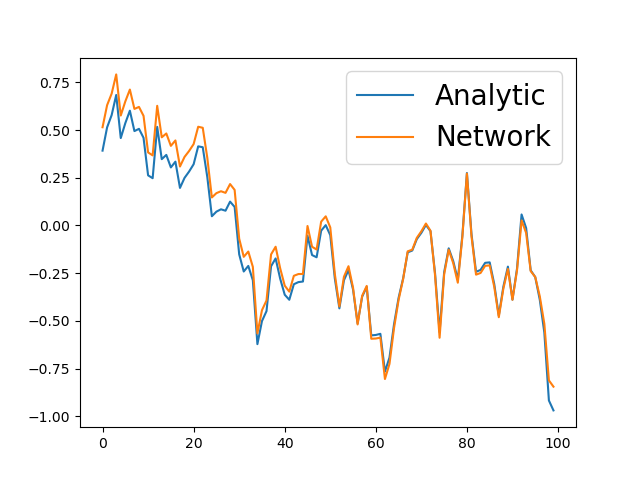}
\caption*{Trajectory 1}
\end{minipage}
\begin{minipage}[t]{0.32\linewidth}
  \centering
 \includegraphics[width=\textwidth]{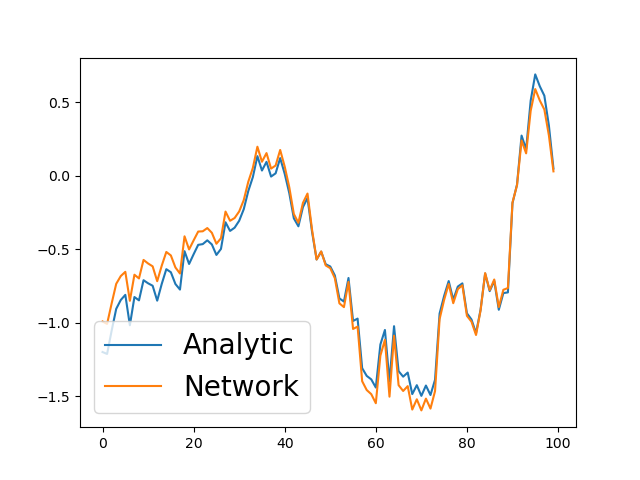}
\caption*{Trajectory 2}
\end{minipage}
\begin{minipage}[t]{0.32\linewidth}
  \centering
 \includegraphics[width=\textwidth]{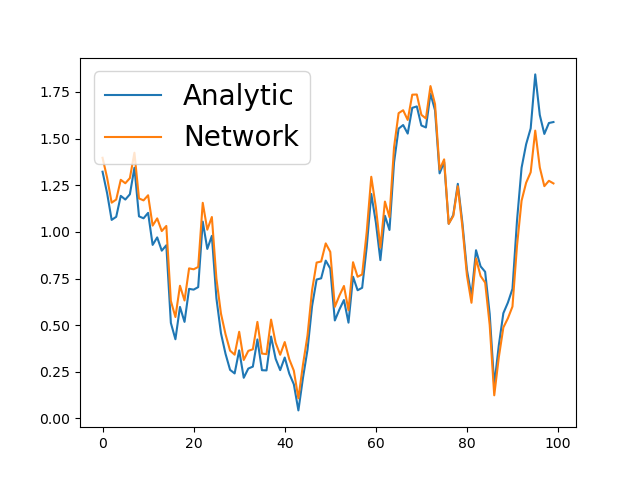}
\caption*{Trajectory 3}
\end{minipage}
      \caption{Trajectories of control with  $\upsilon_0^2=0.9$ \label{fig:trajSys}}
 \end{figure}

On Figure \ref{fig:trajSys},  we plot 3 trajectories of the optimal control and the ones calculated with moment neural networks,  
and we observe that the control is very well estimated.

In this LQ example, we know that the suitable  number of moments to take is $L=2$. 
In the real test case, we do not know which $L$ to take. In  Table \ref{tab:sysMom4}, we give the results with $L=4$, which shows that the results are also very accurate and do not depend on $L$ being small.

\begin{table}[H]
 \centering
    \begin{tabular}{|c|c|c|c|c|c|} \hline 
       $\upsilon_0^2$  &  0. & 0.1 & 0.2 & 0.3 & 0.4  \\ \hline 
         Anal & 0.3869 & 0.4095 & 0.4320 &  0.4546 & 0.4771 \\ \hline 
   Calc      
         & 0.3917 &0.4159 &0.4376 &0.4588 &0.4801 \\ \hline
MSE & 0.0000 &0.0000 &0.0001 &0.0001 &0.0001 \\ \hline
    \end{tabular} 
 \begin{tabular}{|c|c|c|c|c|c|c|} \hline 
       $\upsilon_0^2$  &   0.5 & 0.6 & 0.7 &0.8 & 0.9 \\ \hline 
         Anal &   0.4997 & 0.5222&  0.5448 & 0.5674 & 0.5899\\ \hline 
   Calc      
         & 0.5023 &0.5249 &0.5471 &0.5688 &0.5891 \\ \hline
MSE & 0.0000 &0.0001 &0.0001 &0.0002 &0.0003 \\ \hline
    \end{tabular}
    \caption{\footnotesize{Results for the systemic case using $L=4$, $\rho^A=0.0005$, $\rho^C=0.01$. Training time is 115183s}.
    \label{tab:sysMom4}}
\end{table}

 \subsubsection{An optimal trading  example}
 
 We consider an optimal trading model  taken from \cite{frietal23}: 
\begin{equation}  \label{modeltrading}
\begin{cases}
b(x,\mu,a) &= \;  a, \quad  \sigma \mbox{ positive constant }  \\
f(x,\mu,a)  &  = \;  a^2  + 2Pa, \quad  \quad  g(x,\mu) \; = \;  \gamma(x-\bar\mu)^2, 
\end{cases}
\end{equation}
for $(x,\mu,a)$ $\in$ $\R\times\Pc_2(\R)\times\R$,  with $P$  $>$ $0$ the constant transaction price per trading, and $\gamma$ $>$ $0$ the risk aversion parameter.

In this LQ  framework, the  value function has the form as in  \eqref{eq:sys1D} with
\beqs
K(t)   \; = \;  \frac{\gamma}{1 + \gamma(T-t)},  & &   R(t) \; = \;  \sigma^2 \log (1 + \gamma(T-t)) - P^2 (T-t), 
\enqs
while the optimal control is given by 
\begin{align}
\alpha_t^* &= \; - K(t) (X_t  - \E[X_t]) - P, \quad  0 \leq t \leq T. 
\end{align}

We take the following parameters  : $P=3, \gamma=3, \sigma = 1, T =0.5$, and  $n=100$, $M=10000$. 
 At each gradient iteration, the initial distribution is sampled with
\begin{align*}
X_0  &\sim \; \mu_0 =   \bar\mu_0  + \upsilon_0 {\mathcal N}(0,1), 
\end{align*}
where $(\bar\mu_0 ,\upsilon_0^2)$  are sampled from  $\left(0.4 {\cal U}([0,1]) , 0.5 {\cal U}([0,1]) \right)$  for each element of the batch.

\begin{figure}[H]
  \centering
 \begin{minipage}[t]{0.49\linewidth}
  \centering
  \includegraphics[width=\textwidth]{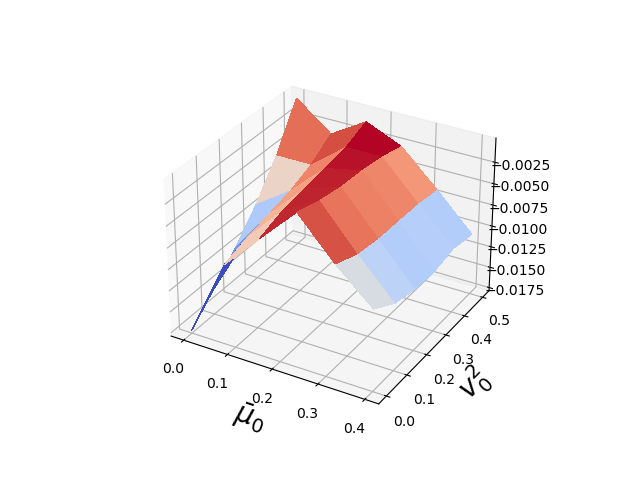}
\caption*{ $L=2$, training time is $110900s$}
\end{minipage}
 \begin{minipage}[t]{0.49\linewidth}
  \centering
  \includegraphics[width=\textwidth]{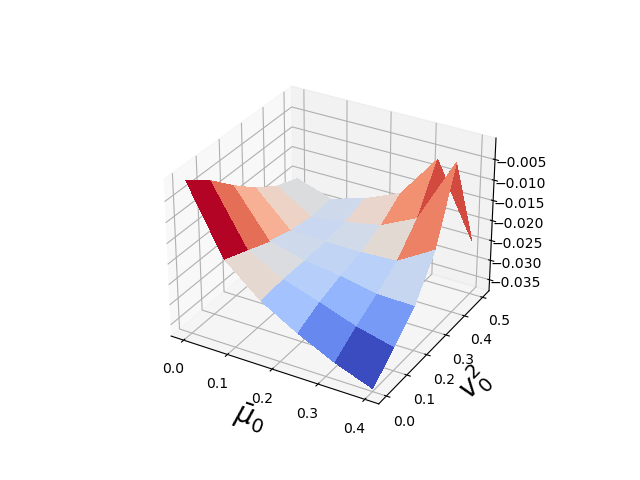}
\caption*{ $L=4$, Training time is $120000s$}
\end{minipage}
\caption{ Relative error  with  $\rho^A=0.0005$, $\rho^C=0.01$, 
$\hat N = 9000$.  \label{fig:tradingErr}}
\end{figure}

\begin{figure}[H]
     \centering
 \begin{minipage}[t]{0.32\linewidth}
  \centering
 \includegraphics[width=\textwidth]{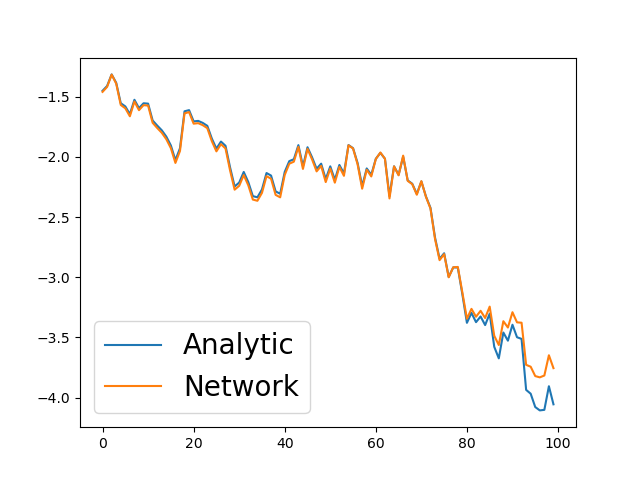}
\caption*{Trajectory 1}
\end{minipage}
\begin{minipage}[t]{0.32\linewidth}
  \centering
 \includegraphics[width=\textwidth]{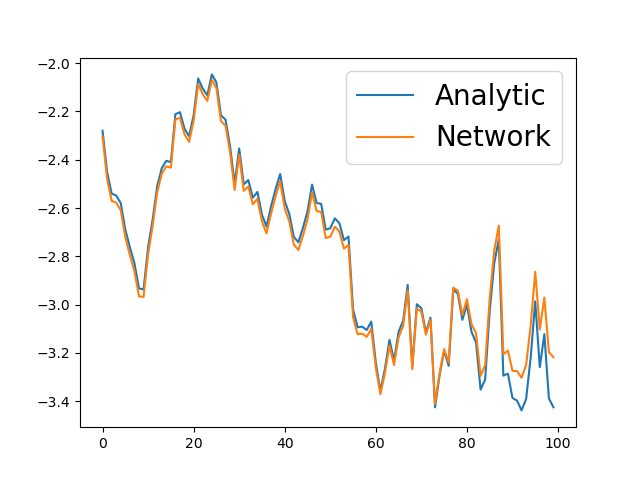}
\caption*{Trajectory 2}
\end{minipage}
\begin{minipage}[t]{0.32\linewidth}
  \centering
 \includegraphics[width=\textwidth]{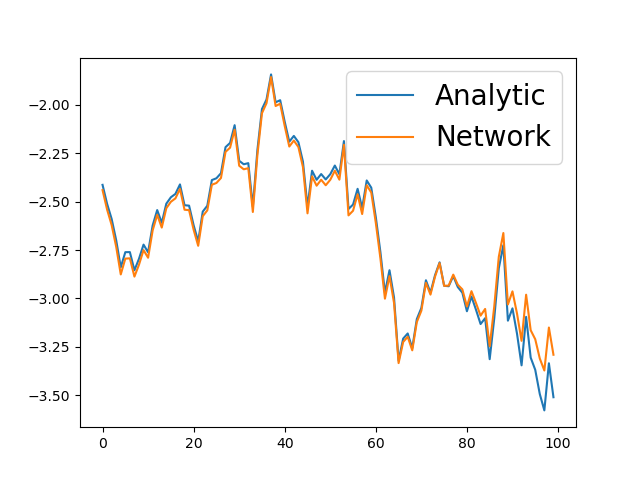}
\caption*{Trajectory 3}
\end{minipage}
      \caption{Controlled trajectories  for $\bar\mu_0=0.16$, $\upsilon_0^2=0.1$, $\rho^A=0.0005$, $\rho^C=0.01$, $L=2$.\label{fig:tradingTraj}}
 \end{figure}

The relative error  is plotted on  Figure \ref{fig:tradingErr} by varying  $(\bar\mu_0,\upsilon_0)$, 
while the trajectories of the control (optimal vs moment neural netwok)  are plotted on Figure \ref{fig:tradingTraj}.  Again, in this LQ example, the suitable  number of moments to take is $L=2$, and when we increase $L$, convergence is more difficult to achieve  and  results become more instable and may depend on the run with the same 
hyper-parameters. Nevertheless we manage to obtain very good results with $L=4$ as shown on Figure \ref{fig:tradingErr}.

\subsubsection{A non linear quadratic mean-field control} \label{sec:NLQ}

We construct an ad-hoc mean-field control model with
\begin{equation}
\begin{cases}
b(t,x,\mu,a) &= \; \beta(t,x,\mu) + a, \quad  \sigma \mbox{ positive constant }   \\
 f(x,\mu,a)  & = F(t,x,\mu) + \frac{1}{2} |a|^2, \quad   g(x, \mu) \;  =  \E_{\xi\sim\mu}[w(x-\xi)]
 \end{cases}
\end{equation}
for some smooth $C^2$ even function $w$ on $\R$, e.g. $w(x)$ $=$ $\cos(x)$, and $F$ is a function to be chosen later.  

In this case, the  optimal feedback control   valued in $A$ $=$ $\R$ is given by
\beqs
\mfa^\star(t,x,\mu) &=& \hat\mra(t,x,\Uc(t,x,\mu)) \; = \;  - \Uc(t,x,\mu) \; = \; -  \partial_\mu v(t,\mu)(x)  \;\; \\
 & & \mbox{ with } v(t,\mu) \; = \;  \E_{\xi\sim\mu}[V(t,\xi,\mu)],
\enqs
and  $V$ is solution to the Master Bellman equation (see section 6.5.2 in \cite{cardel18}): 
\begin{align}
\partial_t V(t,x,\mu) + \big( \beta(t,x,\mu) - \Uc(t,x,\mu) \big)  \partial_x V(t,x,\mu)  +   \frac{\sigma^2}{2}  \partial_{xx}^2 V(t,x,\mu) & \nonumber \\
+ \;  \E_{\xi\sim\mu} \Big[ \big( \beta(t,\xi,\mu) - \Uc(t,\xi,\mu) \big)  \partial_\mu V(t,x,\mu)(\xi)   +   \frac{\sigma^2}{2}  \partial_{x'}\partial_\mu V(t,x,\mu)(\xi)  \Big] & \nonumber \\
\quad + \;   F(t,x,\mu) + \frac{1}{2}   | \Uc(t,x,\mu)|^2   & = \; 0, \label{HJB} 
\end{align} 
with the terminal condition $V(T,x,\mu)$ $=$ $g(x,\mu)$. 

We look for a solution to the Master equation in the form: $V(t,x,\mu)$ $=$ $e^{T-t}\E_{\xi\sim\mu}[w(x-\xi)]$.  For such function $V$, we have $\partial_t V(t,x,\mu)$ $=$ $-V$, 
\beqs
\partial_x V(t,x,\mu) &=& e^{T-t} \E_{\xi\sim\mu} [ w'(x-\xi) ], \quad   \partial_{xx}^2 V(t,x,\mu) \; = \;  e^{T-t} \E_{\xi\sim\mu} [ w''(x-\xi) ] \\
\partial_\mu V(t,x,\mu)(\xi) &=& -  e^{T-t}  w'(x-\xi), \quad \partial_{x'}\partial_\mu V(t,x,\mu)(\xi) \; = \; e^{T-t}  w''(x-\xi), 
\enqs
and \
\beqs
\Uc(t,x,\mu) &=& e^{T-t} \E_{\xi\sim\mu} [ w'(x-\xi) - w'(\xi- x) ] \; = \; 2  e^{T-t} \E_{\xi\sim\mu} [ w'(x-\xi) ]  \; = \; 2 \partial_x V(t,x,\mu). 
\enqs
since $w$ is even. By plugging these derivatives expressions of $V$  into the l.h.s. of \eqref{HJB}, we then see that by choosing $F$ equal to 
\beqs
F(t,x,\mu) &=& e^{T-t} \E_{\xi\sim\mu} \Big[ (w- \sigma^2 w'')(x-\xi) + ( \beta(t,\xi,\mu) - \beta(t,x,\mu)) w'(x-\xi) \Big] \\
& & \; -  \;  2 e^{2(T-t)}  \E_{(\xi,\xi')\sim\mu\otimes\mu} \big[ w'(x-\xi)  w'(\xi- \xi') \big],
\enqs
the function $V$ satisfies the Master Bellman equation, hence is the value function to the mean-field control problem.

Actually, with the choice of $w(x)$ $=$ $\cos(x)$, and using trigonometric relations, we have
{\footnotesize
\beqs
F(t,x,\mu) &=& \cos(x)  \Big[ e^{T-t} \left( (1 + \sigma^2) \E_{\xi\sim\mu}[\cos(\xi)] + \E_{\xi\sim\mu}[ \sin(\xi) \beta(t,\xi, \mu)] - \beta(t,x, \mu) \E_{\xi\sim\mu}[\sin(\xi)] \right)  \\
& & \quad - \;  2 e^{2(T-t)} \left( \E_{\xi\sim\mu}[ \sin(\xi) \cos(\xi)]  \E_{\xi\sim\mu}[\sin(\xi)]  -\E_{\xi\sim\mu}[\sin^2(\xi)] \E_{\xi\sim\mu}[\cos(\xi)] \right) \Big]  \\
&& \quad + \; \sin(x) \Big[ e^{T-t}  \left(  (1 + \sigma^2)  \E_{\xi\sim\mu}[\sin(\xi)]  -\E_{\xi\sim\mu}[ \beta(t,\xi, \mu) \cos(\xi)]  +\beta(t,x ,\mu) \E_{\xi\sim\mu}[\cos(\xi)]  \right)  \\
& & \quad - \;  2e ^{2(T-t)} \left( \E_{\xi\sim\mu}[\sin(\xi)\cos(\xi)]  \E_{\xi\sim\mu}[\cos(\xi) ]  -  \E_{\xi\sim\mu}[\cos^2(\xi)]  \E_{\xi\sim\mu} [\sin(\xi)]  \right) \Big]. 
\enqs
}

\begin{figure}[H]
     \centering
      \begin{minipage}[t]{0.49\linewidth}
  \centering
 \includegraphics[width=\textwidth]{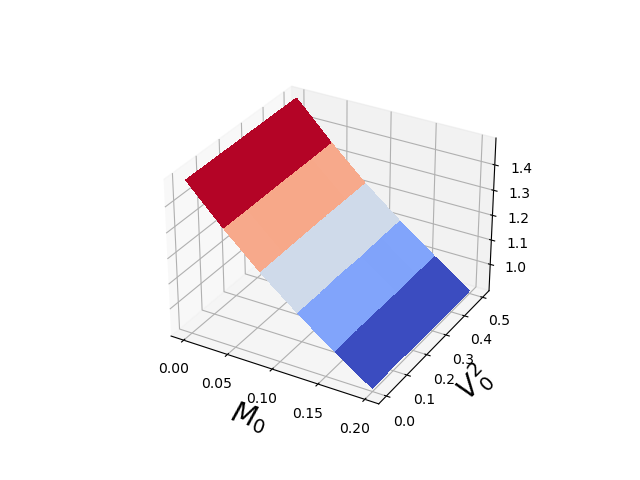}
\caption*{Analytic value}
\end{minipage}
 \begin{minipage}[t]{0.49\linewidth}
  \centering
 \includegraphics[width=\textwidth]{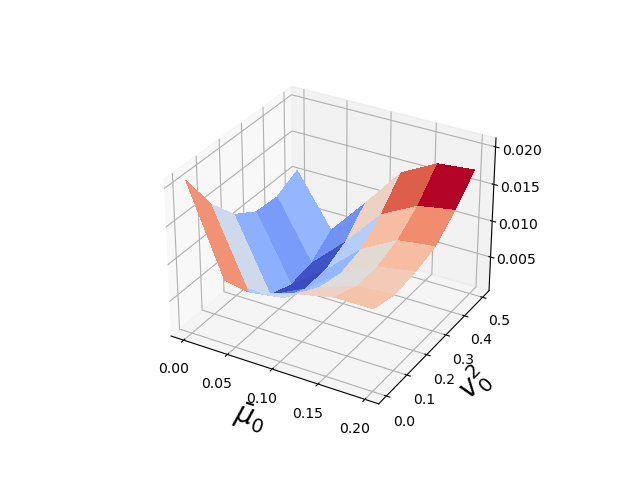}
\caption*{Relative error}
\end{minipage}
\caption{\footnotesize{Non LQ model  using $\hat N =9000$ gradient iterations, $L=3$,  $\rho^A=0.0005$, $\rho^C=0.02$.  Training time is $45200$s.} \label{fig:NLQErr}}
\end{figure}

For the test case, we take
\begin{align}
\beta(t, x , \mu) & = \;   \kappa ( \bar\mu  -x),
\label{eq:trendNLQ}
\end{align}
with the parameters  $\kappa = \sigma =1$, $T=0.4$, $n=40$, $M=20000$.   At each gradient iteration, the initial distribution is sampled with
\begin{align*}
X_0  & \sim \; \mu_0 =  \bar\mu_0  + \upsilon_0 {\mathcal N}(0,1)
\end{align*}
where $(\bar\mu_0 ,\upsilon_0^2)$ is sampled from  $\left(0.2 {\cal U}([0,1]) , 0.5 {\cal U}([0,1]) \right)$ for each element of the batch.  
On Figure \ref{fig:NLQErr}, we give the analytic solution depending on $(\bar\mu_0,\upsilon_0^2)$ and the relative error obtained by the algorithm. We observe that the results are quite accurate.

\begin{figure}[H]
\begin{minipage}[t]{0.32\linewidth}
  \centering
 \includegraphics[width=\textwidth]{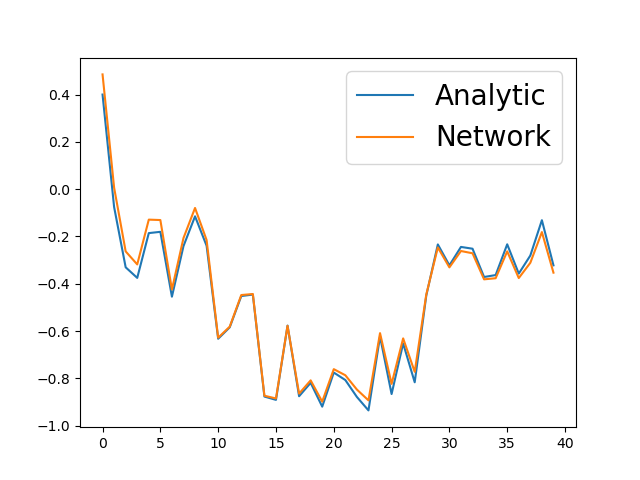}
\caption*{Trajectory 1}
\end{minipage}
\begin{minipage}[t]{0.32\linewidth}
  \centering
 \includegraphics[width=\textwidth]{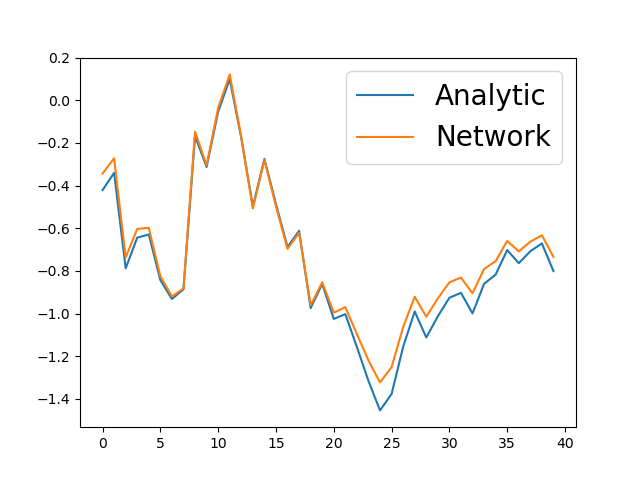}
\caption*{Trajectory 2}
\end{minipage}
\begin{minipage}[t]{0.32\linewidth}
  \centering
 \includegraphics[width=\textwidth]{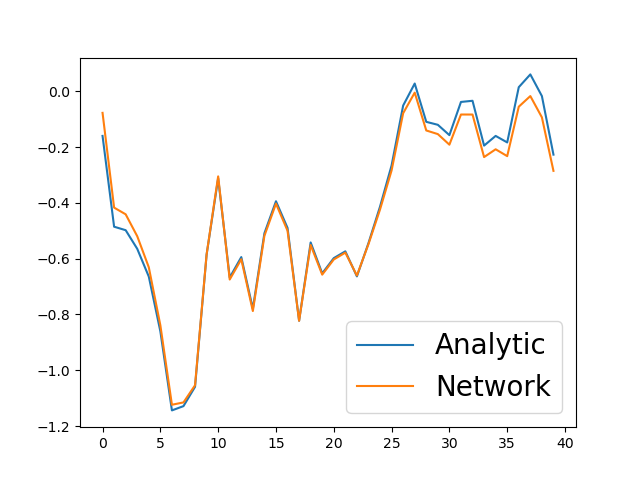}
\caption*{Trajectory 3 }
\end{minipage}
      \caption{\footnotesize{Control trajectories in a non LQ model using $\hat N =9000$ gradient iterations, $L=3$,  $\rho^A=0.0005$, $\rho^C=0.02$, for $(\bar\mu_0, \upsilon_0^2)= (0.16, 0.1)$.} \label{fig:NLQTraj}}
 \end{figure}

We plot in Figure \ref{fig:NLQTraj} the trajectories of the optimal control vs the moment neural network, and observe that they are very close.

\begin{figure}[H]
     \centering
 \begin{minipage}[t]{0.47\linewidth}
  \centering
 \includegraphics[width=\textwidth]{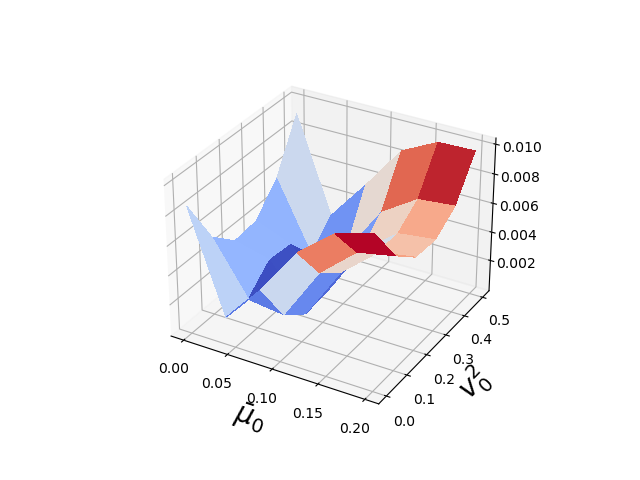}
\caption*{$L =2$}
\end{minipage}
\begin{minipage}[t]{0.47\linewidth}
  \centering
 \includegraphics[width=\textwidth]{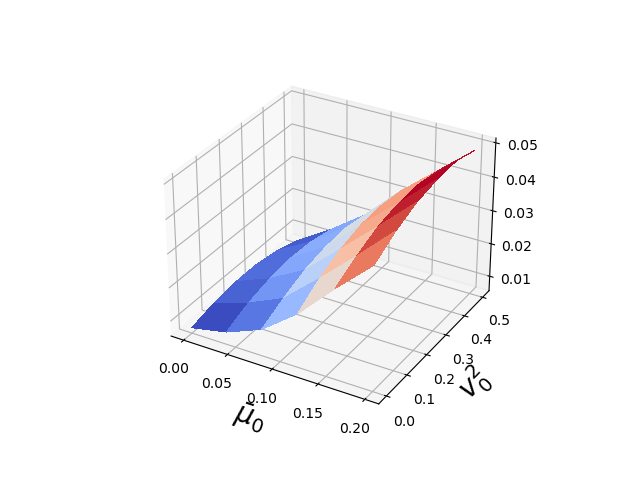}
\caption*{ $L =4$}
\end{minipage}
\begin{minipage}[t]{0.47\linewidth}
  \centering
 \includegraphics[width=\textwidth]{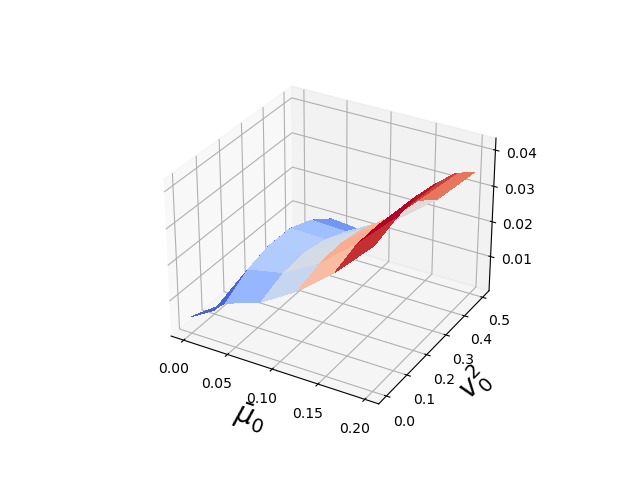}
\caption*{$L =5$} 
\end{minipage}
\begin{minipage}[t]{0.47\linewidth}
  \centering
\includegraphics[width=\textwidth]{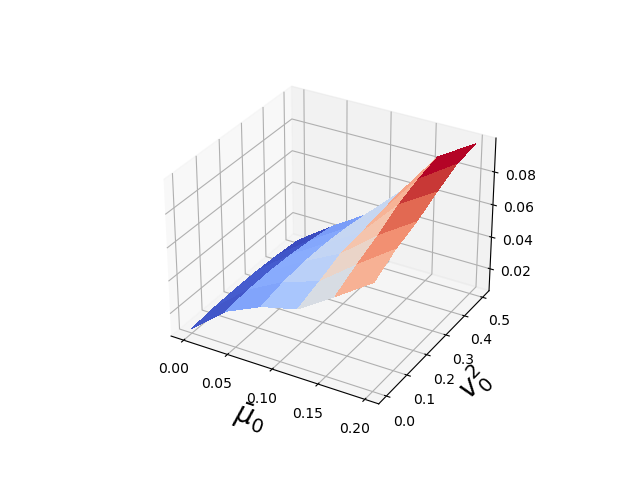}
\caption*{ $L =6$}
\end{minipage}
      \caption{\footnotesize{Relative error in a non LQ model  for different values of $L$.  $\hat N =9000$ gradient iterations,  $\rho^A=0.0005$, $\rho^C=0.02$. } \label{fig:NLQMom}}
 \end{figure}

Figure \ref{fig:NLQMom} shows that using $L=2$ or $3$ is optimal in terms of relative error,  while the convergence is more difficult to achieve for high values of $L$.

 \subsubsection{A multi-dimensional LQ example}

 We consider a multi-dimensional extension of the LQ systemic risk model in  Section \ref{sec:sys1D}, by supposing that on each dimension, the dynamics satisfies the same equation with independent Brownian motions,  and that the cost functions are 
 the sum over each component of the cost function in the univariate case.  In this case, the value function  is given by  $V(t,x,\mu)$ $=$ $\sum_{i=1}^d V_1(t,x_i,\mu_i)$,  for $t$ $\in$ $[0,T]$, $x$ $=$ $(x_i)_{i\in\llbracket 1,d\rrbracket}$ $\in$ $\R^d$, $\mu_i$ is the $i$-th marginal 
 law of $\mu$ $\in$ $\Pc_2(\R^d)$, and $V_1$ is the value function in the univariate model given by \eqref{eq:sys1D}. 
 
We keep the same parameters as in  Section \ref{sec:sys1D}, with  a number of time steps $n=50$,  $M= 10000$, $L=2$, and test in dimension $d=2$  and $3$. 
At each gradient iteration, the initial distribution is sampled from  
\begin{align*}
X_0 &  \sim \;  \mu_0 =   {\mathcal N}(0, \upsilon_0), 
\end{align*}
where $\upsilon_0$ is the diagonal $d\times d$-matrix with diagonal elements $\upsilon_{0,i}$ sampled from $u_i {\cal U}([0,1])$, with constants $u_i$ $\in$ $[0,1]$, $i$ $=$ $1,\ldots,d$,   for each element of the batch.

\begin{figure}[H]
     \centering
\begin{minipage}[t]{0.48\linewidth}
  \centering
 \includegraphics[width=\textwidth]{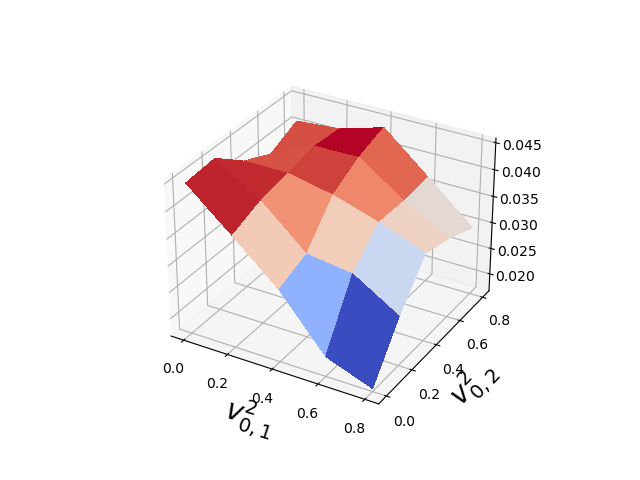}
      \caption*{ $L=2$, $\hat N =6000$ ,  $\rho^A=0.001$}
 \end{minipage}
 \begin{minipage}[t]{0.48\linewidth}
  \centering
 \includegraphics[width=\textwidth]{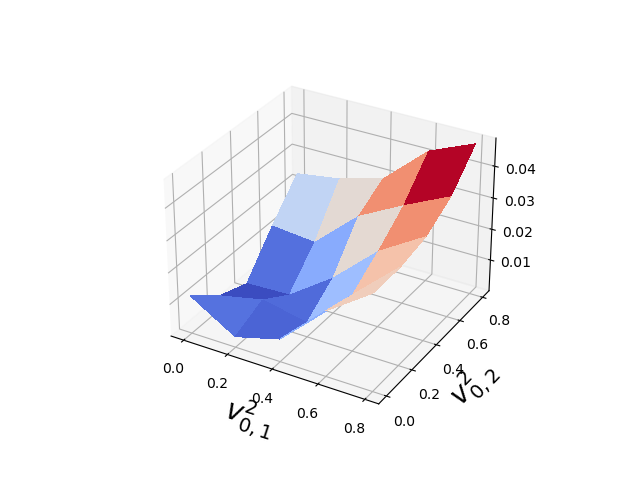}
      \caption*{ $L=4$, $\hat N =9000$ ,  $\rho^A=0.0005$}
 \end{minipage}
   \caption{\footnotesize{Relative error in dimension  $d$ $=$ $2$,  with  $\rho^C=0.01$.}  \label{fig:2D}}
 \end{figure}

\begin{figure}[H]
     \centering
 \includegraphics[width=0.5\textwidth]{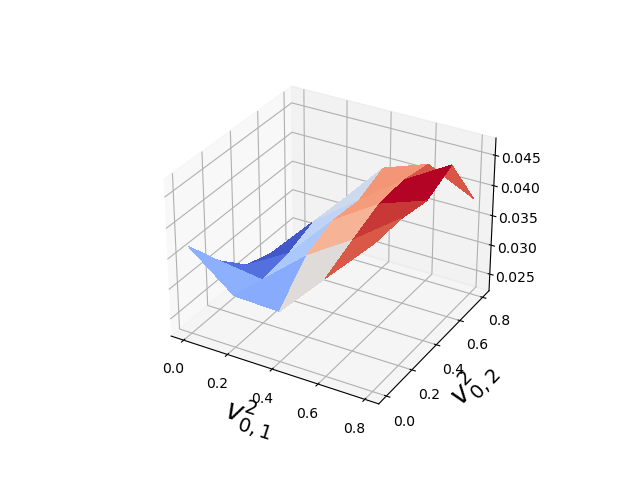}
      \caption{\footnotesize{Relative error for $L=2$ in  dimension $d$ $=$ $3$, with $\hat N =9000$ gradient iterations,  $\rho^A=0.0005$, $\rho^C=0.01$. Training time is 144000s}. \label{fig:3D}}
 \end{figure}

We plot in Figure \ref{fig:2D} the relative error in dimension $d$ $=$ $2$ by varying $(\upsilon_{0,1},\upsilon_{0,2})$, and for $L$ $=$ $2$ and $L$ $=$ $4$.  In Figure \ref{fig:3D}, we plot  the relative error in dimension $d$ $=$ $3$ by varying $(\upsilon_{0,1},\upsilon_{0,2})$, and for $\upsilon_{0,3}$ $=$ $0$.

\subsection{A  non LQ  example with controlled volatility}

We consider a one-dimensional model with 
\begin{equation}
\begin{cases}
b(t,x,\mu,a) & = \; \beta(t,x,\mu), \quad \sigma(t,x,\mu,a) \; = \; a, \\
f(x,\mu,a) &= \; F(t,x,\mu) + \frac{1}{2} P |a|^2  - a, \quad g(x, \mu) \; = \;   \E_{\xi\sim\mu}[w(x-\xi)],
\end{cases}
\end{equation}
where $P$ is a positive constant,  $w$ is a smooth $C^2$ even function on $\R$, e.g. $w(x)$ $=$ $\cos(x)$, and $F$ is a function to be chosen later.  Notice that  $C_\theta$ $\equiv$ $0$, and 
\beqs
\vartheta_\theta(t,x,\mu) &=& m_\theta(t,x,\mu)^2 + \lambda, 
\enqs
where $\lambda$ $=$ $\lambda(e)$ (depending on the epoch $e$) is the exploration parameter of $\pi_\theta(.|t,x,\mu)$ $=$ $\Nc(m_\theta(t,x,\mu),\lambda)$.

In this model, the  optimal feedback control   valued in $A$ $=$ $\R$ is given by
\begin{align}
\mfa^\star(t,x,\mu) &= \;  \hat\mra(t,x, \partial_x \Uc(t,x,\mu)) \; = \;  
\frac{ 1}{ P + \partial_x\Uc(t,x,\mu)}, \\
& \mbox{ with } \;\;  \Uc(t,x,\mu) \; = \; \partial_\mu v(t,\mu)(x), \quad v(t,\mu) \; = \; \E_{\xi\sim\mu}[V(t,\xi,\mu)], 
\end{align}
and  $V$ is solution to the Master Bellman equation: 
\begin{align}
\partial_t V(t,x,\mu) +  \beta(t,x,\mu)  \partial_x V(t,x,\mu)  +   \frac{1}{2} \frac{1}{(P + \partial_x\Uc(t,x,\mu))^2}  \partial_{xx}^2 V(t,x,\mu) & \nonumber \\
+ \;  \E_{\xi\sim\mu} \Big[  \beta(t,\xi,\mu)  \partial_\mu V(t,x,\mu)(\xi)   +  \frac{1}{2} \frac{1}{(P + \partial_x\Uc(t,\xi,\mu))^2}    \partial_{x'}\partial_\mu V(t,x,\mu)(\xi)  \Big] & \nonumber \\
\quad + \;   F(t,x,\mu) + \frac{1}{2} \frac{ P}{ (P + \partial_x\Uc(t,x,\mu))^2}  - \frac{1}{P + \partial_x\Uc(t,x,\mu)}  & = \; 0, \label{HJB} 
\end{align} 
with the terminal condition $V(T,x,\mu)$ $=$ $g(x,\mu)$.

We look for a solution to the Master equation in the form: $V(t,x,\mu)$ $=$ $e^{T-t}\E_{\xi\sim\mu}[w(x-\xi)]$, and by similar calculations as in Section \ref{sec:NLQ}, we would have $\Uc(t,x,\mu)$ 
$=$  $2 \partial_x V(t,x,\mu)$ $=$ $2e^{T-t} \E_{\xi\sim\mu} [ w'(x-\xi)]$.  Therefore, with $w(x)$ $=$ $\cos(x)$, and by choosing $F$ equal to  
\begin{align*}
 F(t,x,\mu) = &-  \frac{  P}{2 (P - 2 e^{T-t} \E_{\xi\sim\mu}[\cos(x-\xi)] )^2}  + \frac{1}{P - 2 e^{T-t} \E_{\xi\sim\mu}[\cos(x-\xi)]} + \\
 & \E_{\xi\sim\mu}[\cos(x-\xi)] e^{T-t} (1+ \frac{1}{2} \frac{1}{(P -2 e^{T-t} \E_{\xi\sim\mu}[\cos(x-\xi)] )^2})+ \\
 & e^{T-t} \E_{\xi\sim\mu}[ (\beta(t,x, \mu) -\beta(t,\xi, \mu)) \sin(x-\xi)] + \\
 & e^{T-t} \E_{\xi\sim\mu}[ \cos(x- \xi) \frac{1}{2}  \frac{1}{(P -2 e^{T-t} \E_{\xi'\sim\mu}[\cos(\xi-\xi')] )^2} ],
\end{align*}
the function $V$ satisfies the Master Bellman equation, hence is the value function to the mean-field control problem. 

To be easily computable, the function $F$  can be rewritten  using trigonometric relations as 
\begin{align*}
F(t,x,\mu)= & - \frac{P}{2}\frac{  1}{ (P - 2 e^{T-t} [\cos(x) \overline\cos_\mu + \sin(x) \overline\sin_\mu]  )^2} +  \frac{  1}{ P - 2 e^{T-t} [\cos(x) \overline\cos_\mu + \sin(x) \overline\sin_\mu]  }  \\
& \quad + \;  e^{T-t} [ \cos(x) \overline\cos_\mu + \sin(x) \overline\sin_\mu ] (1+ \frac{1}{2} \frac{1}{(P -2 e^{T-t} [ \cos(x) \overline\cos_\mu + \sin(x) \overline\sin_\mu ]  )^2}) \\
& \quad + \;  e^{T-t} [ \beta(t,x, \mu) \sin(x) \overline\cos_\mu -\beta(t,x, \mu) \cos(x) \overline\sin_\mu \\
& \quad  \quad - \; \E_{\xi\sim\mu}[\beta(t,\xi, \mu) \cos(\xi)] \sin(x) + \E_{\xi\sim\mu}[\beta(t,\xi, \mu) \sin(\xi)] \cos(x)]  \\
& \quad + \; \frac{e^{T-t}}{2}   \cos(x) \E_{\xi\sim\mu}[ \frac{\cos( \xi) }{2[P -2 e^{T-t}( \cos(\xi) \overline\cos_\mu  + \sin(\xi) \overline\sin_\mu)]^2}]  \\
& \quad + \;  \frac{e^{T-t}}{2}   \sin(x) \E_{\xi\sim\mu}[ \frac{\sin( \xi) }{2[P -2 e^{T-t}( \cos(\xi) \overline\cos_\mu  + \sin(\xi) \overline\sin_\mu)]^2}] 
\end{align*}
with the notations: $\overline \cos_\mu$ $:=$ $\E_{\xi \sim \mu}[\cos( \xi)]$,  $\overline\sin_\mu$ $:=$  $\E_{\xi \sim \mu}[\sin( \xi)]$.

We take $P = 2.2 e^{T}$ so  that the control is bounded,  the same trend  $\beta$ as in  \eqref{eq:trendNLQ},  and  parameters as in section \ref{sec:NLQ}: $\kappa = \sigma =1$, $T=0.4$, $n=40$, $M=20000$.  At each gradient iteration, the initial distribution is sampled with
\begin{align*}
x_0  &\sim \;  \mu_0 =   \bar\mu_0  + \upsilon_0 {\mathcal N}(0,1)
\end{align*}
where $(\bar\mu_0 ,\upsilon_0^2)$ is sampled from $\left(0.2 U([0,1]) , 0.5 U([0,1]) \right)$.   Controlling the volatility is more difficult than controlling the trend, and it is crucial for the method that $\rho^A$ is very small. We take $\hat N=9000$ gradient iterations. 
Training time with $L=3$ is $67228$s, while it takes $69073$s  for $L=4$.

\begin{figure}[H]
     \centering
 \begin{minipage}[t]{0.49\linewidth}
  \centering
 \includegraphics[width=\textwidth]{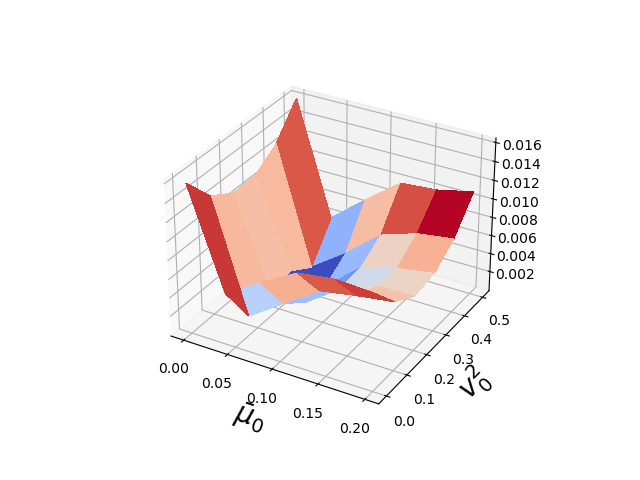}
\caption*{$L =3$, $\rho^A=0.0005$,  $\rho^C=0.025$}
\end{minipage}
\begin{minipage}[t]{0.49\linewidth}
  \centering
 \includegraphics[width=\textwidth]{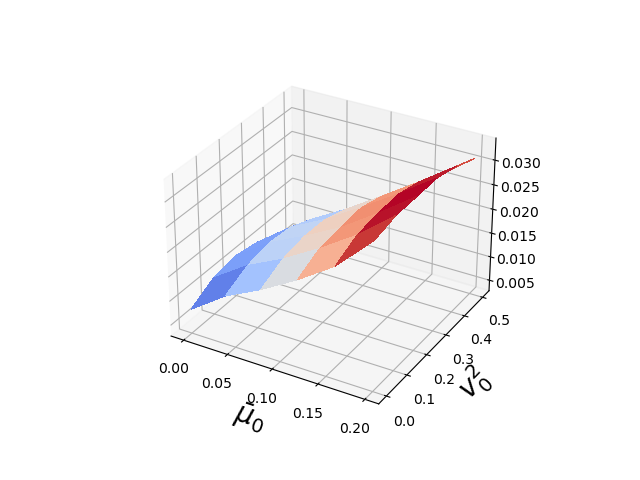}
\caption*{$L =4$, $\rho^A=0.0002$ and $\rho^C=0.01$}
\end{minipage}
\caption{\footnotesize{Relative error in a non LQ model with controlled volatility}. \label{fig:NLQVolErr}}
\end{figure}

\begin{figure}[H]
     \centering
\begin{minipage}[t]{0.32\linewidth}
  \centering
 \includegraphics[width=\textwidth]{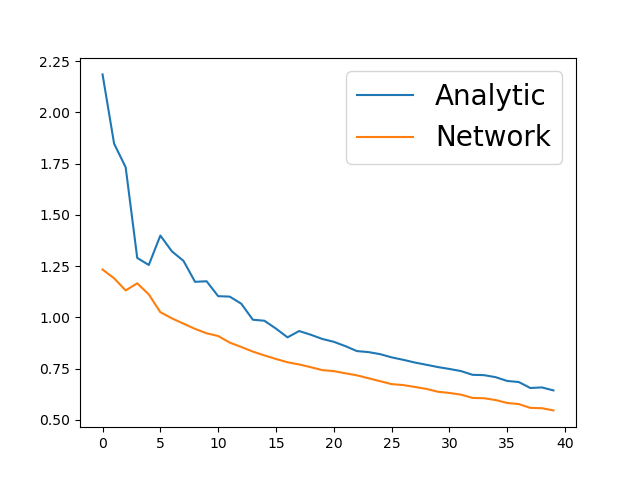}
\caption*{Trajectory 1}
\end{minipage}
\begin{minipage}[t]{0.32\linewidth}
  \centering
 \includegraphics[width=\textwidth]{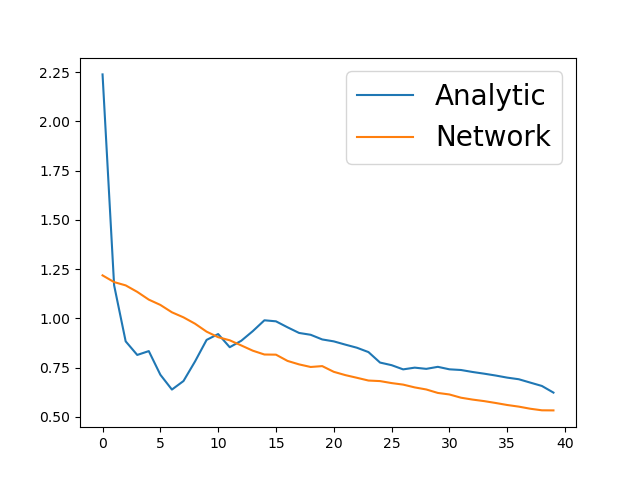}
\caption*{Trajectory 2}
\end{minipage}
\begin{minipage}[t]{0.32\linewidth}
  \centering
 \includegraphics[width=\textwidth]{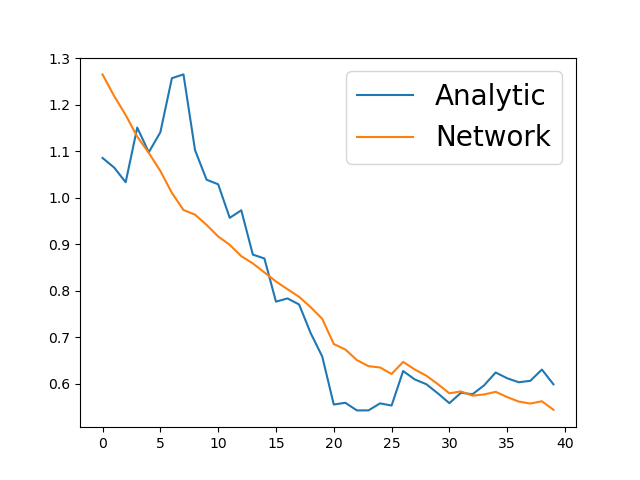}
\caption*{Trajectory 3}
\end{minipage}
      \caption{\footnotesize{Control trajectories in a non LQ model with controlled volatility, using $L= 3$,  with  $(\bar\mu_0, \upsilon_0^2)= (0.04, 0.4)$}.  \label{fig:NLQVolTraj}}
 \end{figure}

On Figure \ref{fig:NLQVolErr},  we give the relative error obtained with $L=3$ and $L=4$. Notice that with $L=3$, $\rho^A$ is small and that we have to take $ \rho^A$ even smaller with $L=4$.
The relative error is small, but the control is not as well approximated as in the controlled drift example, as shown in Figure \ref{fig:NLQVolTraj}.

\section{Conclusion}

In this study, we have presented a robust  effective resolution to the challenging problem of mean-field control within a partially model-free continuous-time framework. Leveraging policy gradient techniques and actor-critic algorithms, our approach has demonstrated the valuable role of moment neural networks in the sampling 
of distributions. We have illustrated the significance of maintaining a low number of moments (typically two or three) while underscoring the critical role played by fine-tuning learning rates for actor and critic updates. 
 
 Subsequent developments could encompass the extension to  non-separable forms within the state and control components of drift and diffusion coefficients. Furthermore, a compelling direction for further investigation could involve mean-field dynamics governed by jump diffusion processes, where the intensities of the jumps remain unknown.

\printbibliography

@article{jiazho21,
	author = {Jia, Yanwei and Zhou, Xun Yu},
	journal = {to appear in {Journal of Machine Learning Research}},
	title = {Policy gradient and actor critic learning in continuous time and space: theory and algorithms},
	year = {2021}}

@book{Sutbar18,
	author = {Sutton, Richard and Barto, Andrew},
	publisher = {Cambridge, MA:MIT},
	title = {Reinforcement learning: an introduction},
	year = {2018, 2nd edition}}

@article{guetal21,
	author = {Gu, Haotian and Guo, Xin and Wei, Xiaoli and Xu, Renyuan},
	journal = {SIAM Journal on Mathematics of Data Science},
	title = {Mean field controls with {Q}-learning for cooperative {MARL}: convergence and complexity analysis},
	year = {2021}}

@article{carlautan22,
	author = {Carmona, Ren{\'e} and Lauri{\`e}re, Mathieu and Tan, Zongjun},
	journal = {to appear in {Annals of Applied Probability}},
	title = {Model-free mean-field reinforcement learning: mean-field {MDP} and mean-field {Q}-learning},
	year = {2022}}

@article{Ruthotto9183,
	author = {Ruthotto, Lars and Osher, Stanley and Li, Wuchen and Nurbekyan, Levon and Fung, Samy Wu},
	journal = {Proc. Natl. Acad. Sci. USA},
	number = {17},
	pages = {9183-9193},
	title = {A machine learning framework for solving high-dimensional mean field game and mean field control problems},
	volume = {117},
	year = {2020}}

@article{phawar22,
	author = {Pham, Huy{\^e}n and Warin, Xavier},
	journal = {arXiv: 2212.11518},
	title = {Mean-field neural networks-based algorithms for {McKean-Vlasov} control problems},
	year = {2022}}

@article{reistozha21,
	author = {Reisinger, Christoph and Stockinger, Wolfgang and Zhang, Yufei},
	journal = {arXiv: 2108.06740},
	title = {A fast iterative {PDE}-based algorithm for feedback controls of nonsmooth mean-field control problems},
	year = {2021}}

@article{hanhulong22,
	author = {Han, Jiequn and Hu, Ruimeng and Long, Jihao},
	journal = {arXiv: 2204.11924},
	title = {{Learning high-dimensional McKean-Vlasov forward-backward stochastic differential equations with general distribition dependence}},
	year = {2022}}

@article{germikwar22,
	author = {Germain, Maximilien and Mikael, Joseph and Warin, Xavier},
	journal = {Methodology and Computing in Applied Probability},
	title = {{Numerical resolution of McKean-Vlasov FBSDEs using neural networks}},
	year = {2022}}

@article{carlau22,
	author = {Carmona, Ren{\'e} and Lauri\`ere, Mathieu},
	journal = {Annals of Applied Probability},
	number = {6},
	pages = {4065-4105},
	title = {{Convergence analysis of machine learning algorithms for the numerical solution of mean-field control and games: II-the finite horizon case}},
	volume = {32},
	year = {2022}}

@book{cardel18b,
	author = {Carmona, Ren{\'e} and Delarue, Fran{\c c}ois},
	publisher = {Springer},
	title = {Probabilistic Theory of Mean Field Games: vol. II, Mean Field game with common noise and Master equations},
	year = {2018}}

@article{angfoulau21,
	author = {Angiuli, Andrea and Fouque, Jean-Pierre and Lauri\`ere, Mathieu},
	journal = {Mathematics of Control, Signals and Systems},
	pages = {217-271},
	title = {Unified Reinforcement {Q}-Learning for Mean Field Game and Control Problems},
	volume = {34},
	year = {2022}}

@book{cardel18,
	author = {Carmona, Ren{\'e} and Delarue, Fran{\c c}ois},
	publisher = {Springer},
	title = {Probabilistic theory of mean-field games: vol. I, Mean field {FBSDEs}, Control, and Games},
	year = {2018}}

@article{carfousun15,
	author = {Carmona, Ren{\'e} and Fouque, Jean-Pierre and Sun, Li-Hsien},
	journal = {Commun. Math. Sci.},
	number = {4},
	pages = {911-933},
	title = {Mean field games and systemic risk},
	volume = {13},
	year = {2015}}

@article{frietal23,
	author = {Frikha, Noufel and Germain, Maximilien and Lauri\`ere, Mathieu and Pham, Huy{\^e}n and Song, Xuanye},
	journal = {arXiv:2303.06993},
	title = {Actor-critic learning for mean-field control in continuous time},
	year = {2023}}

@article{kingma2014adam,
	author = {Kingma, Diederik P and Ba, Jimmy},
	journal = {arXiv preprint arXiv:1412.6980},
	title = {Adam: A method for stochastic optimization},
	year = {2014}}

@article{warin2023quantile,
	author = {Warin, Xavier},
	journal = {arXiv preprint arXiv:2303.11060},
	title = {Quantile and moment neural networks for learning functionals of distributions},
	year = {2023}}

@article{pham2022mean,
	author = {Pham, Huy{\^e}n and Warin, Xavier},
	journal = {arXiv preprint arXiv:2210.15179, to appear in Neural networks},
	title = {Mean-field neural networks: learning mappings on Wasserstein space},
	year = {2022}}

@article{wang2020reinforcement,
	author = {Wang, Haoran and Zariphopoulou, Thaleia and Zhou, Xun Yu},
	journal = {The Journal of Machine Learning Research},
	number = {1},
	pages = {8145--8178},
	publisher = {JMLRORG},
	title = {Reinforcement learning in continuous time and space: A stochastic control approach},
	volume = {21},
	year = {2020}}

@article{germain2022deepsets,
	author = {Germain, Maximilien and Lauri{\`e}re, Mathieu and Pham, Huy{\^e}n and Warin, Xavier},
	journal = {Journal of Scientific Computing},
	publisher = {Springer},
	title = {DeepSets and their derivative networks for solving symmetric PDEs},
	volume = {91, article 63},
	year = {2022}}

\end{document}